%% file: main.tex
\documentclass[runningheads]{llncs}

\usepackage{eccv}

\usepackage{eccvabbrv}
\usepackage{graphicx}
\usepackage{booktabs}
\usepackage{multirow}

\usepackage{wrapfig}

\usepackage[accsupp]{axessibility}  

\usepackage{hyperref}

\usepackage{orcidlink}

\input{macros}

\usepackage[marginal]{footmisc}

\newcommand{\wlink}[1]{\textcolor{magenta}{{#1}}}

\begin{document}

\title{An Optimization Framework to Enforce Multi-View Consistency for Texturing 3D Meshes} 

\titlerunning{Texturing 3D Meshes with Multi-View Consistency}


\author{Zhengyi Zhao\inst{1}  \and
Chen Song\inst{2} \and
Xiaodong Gu\inst{1} \and 
Yuan Dong\inst{1} \and 
Qi Zuo\inst{1} \and \\
Weihao Yuan\inst{1} \and
Liefeng Bo\inst{1} \and
Zilong Dong\inst{1}\textsuperscript{\dag} \and
Qixing Huang\inst{2}\textsuperscript{\dag} }

\authorrunning{Zhengyi Zhao et al.}


\institute{Institute for Intelligent Computing, Alibaba Group \and
The University of Texas at Austin }


\maketitle


\footnotetext[2]{Corresponding authors:
\wlink{list.dzl@alibaba-inc.com}, \wlink{huangqx@utexas.edu}.}

\input{00_abstract}
\input{01_intro}
\input{02_related}
\input{03_overview}

\input{04_approach}
\input{05_results}

\input{06_limitations} 
\input{07_conclusions}

%
%
\bibliographystyle{splncs04}
\bibliography{main}




\end{document}

%% file: macros.tex
\newcommand{\R}{\mathbb{R}}
\let \bs=\mathbf
\let \set=\mathcal

\def \path {\mathit{path}}

\def \exp {\textup{exp}}

\let \set = \mathcal
\let \bs = \boldsymbol


%% file: 00_abstract.tex
\begin{abstract}
A fundamental problem in the texturing of 3D meshes using pre-trained text-to-image models is to ensure multi-view consistency. State-of-the-art approaches typically use diffusion models to aggregate multi-view inputs, where common issues are the blurriness caused by the averaging operation in the aggregation step or inconsistencies in local features. This paper introduces an optimization framework that proceeds in four stages to achieve multi-view consistency. 
Specifically, the first stage generates an over-complete set of 2D textures from a predefined set of viewpoints using an MV-consistent diffusion process. The second stage selects a subset of views that are mutually consistent while covering the underlying 3D model. We show how to achieve this goal by solving semi-definite programs. The third stage performs non-rigid alignment to align the selected views across overlapping regions. The fourth stage solves an MRF problem to associate each mesh face with a selected view. In particular, the third and fourth stages are iterated, with the cuts obtained in the fourth stage encouraging non-rigid alignment in the third stage to focus on regions close to the cuts. Experimental results show that our approach significantly outperforms baseline approaches both qualitatively and quantitatively. Project page: \href{https://aigc3d.github.io/ConsistenTex}{https://aigc3d.github.io/ConsistenTex}.
\keywords{3D texturing \and Multi-view consistency \and Optimization}
\end{abstract}

%% file: 01_intro.tex
\section{Introduction}

The performance of text-to-image models has reached a state where they can generate photorealistic 2D images of textured objects from a simple text prompt. This ability has stimulated a lot of interest in developing textured mesh models from a text prompt through multiple views of a 3D model. However, the idea of using 2D text-to-image models to synthesize 3D textures faces the fundamental challenge of achieving multi-view consistency among the 2D inputs. 

There are two types of approaches towards achieving multi-view consistency. The first type~\cite{chen2023text2tex,10.1145/3588432.3591503,latent-nerf,Cao_2023_ICCV,Yu_2023_ICCV} generates 3D textures purely based on 2D inputs and performs some type of averaging among overlapping regions of the 2D inputs. Although the goal is to retain high-quality details from the 2D inputs, the results among overlapping regions are usually blurry due to averaging. The second type~\cite{Consistent123,Consistent-1-to-3,Mvdream,SyncDreamer,dmv3d,zero123pp,TextMesh} leverages 3D data to enhance multi-view consistency, e.g., by using attention modules trained from rendered images of 3D models. These methods improved multi-view consistency, but the results tend to overfit rendered images and lose realistic image details in synthetic 2D images.  

\begin{figure}[t]
\centering
\captionsetup{type=figure}
\includegraphics[width=1.0\textwidth]{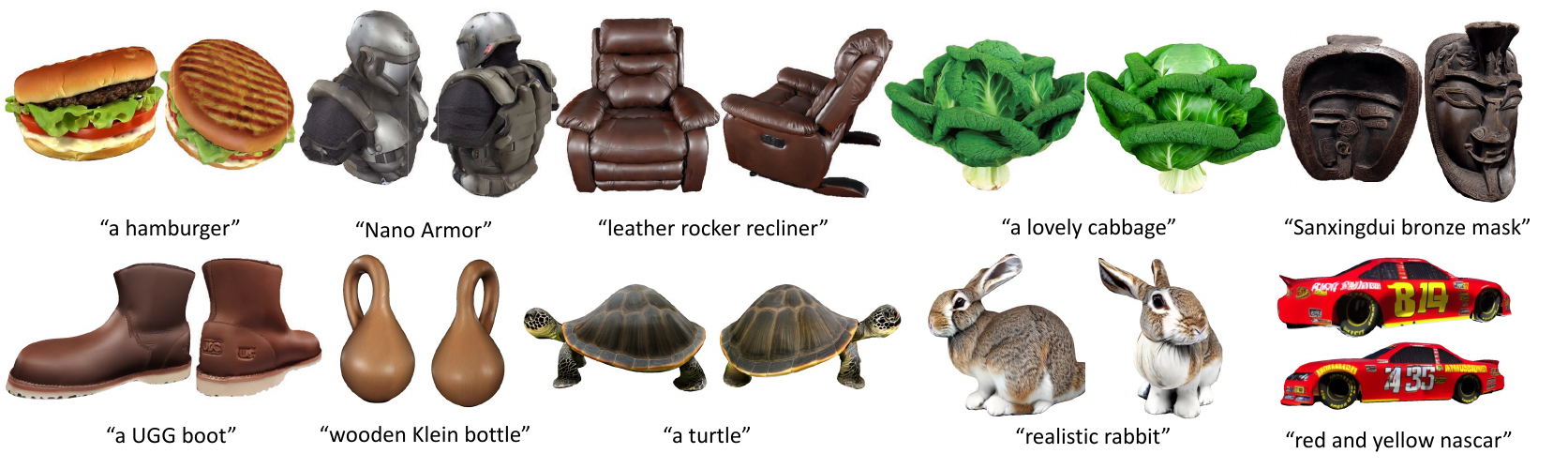}
\captionof{figure}{Given a 3D mesh and a text prompt, we propose an optimization framework to synthesize the multi-view consistent texture. Top: meshes from a 3D generation model. Bottom: artist-created meshes.}
\end{figure}

This paper presents a novel text-to-textured-mesh approach that achieves multi-view consistent results. Our approach makes contributions in both types of methods and combines their strength in a seamless way. While our approach also employs 3D data for training, we focus on promoting multi-view consistency in global style and shape that possess a minimal domain gap between rendered images and synthetic 2D images. On the other hand, we allow for global drifts in image details, which are the main source of overfitting. To achieve multi-view consistency, we perform non-rigid alignments among overlapping regions of synthetic 2D images. Non-rigid alignments are performed among a subset of images selected from an over-complete of synthetic 2D images. 

Specifically, our approach proceeds in four stages. The first stage uses a pre-trained text-to-image model to generate an overcomplete set of candidate textures from a predefined set of viewpoints. The second stage optimizes a subset of views that are mutually consistent. The third stage performs joint alignment among selected 2D textures to maximize consistency across overlapping regions. The fourth stage cuts and stitches the aligned texture images into a consistently textured mesh. The third and fourth stages are iterated in an alternating manner, where the cuts obtained in the fourth stage guide the non-rigid alignment procedure so that alignment focuses on regions near the optimized cuts.   

Experimental results show that our approach outperforms state-of-the-art text-to-texture approaches both qualitatively and quantitatively. Qualitatively, our approach is much more photorealistic and does not present discontinuous or blurry texture elements that are frequent among other approaches. Quantitatively, our approach achieves a better FID score than the existing methods.

%% file: 02_related.tex
\section{Related Work}
\label{Section:Related:Work}
\textbf{Multi-View Representation.} Multi-view representations are used in many 3D recognition approaches~\cite{DBLP:conf/iccv/SuMKL15,DBLP:conf/cvpr/KanezakiMN18,DBLP:conf/cvpr/WeiYS20,DBLP:conf/eccv/KunduYFRBFP20,DBLP:conf/iccv/HamdiGG21,DBLP:conf/iclr/HamdiGG23}, where the key idea is to take advantage of large-scale pre-trained 2D networks. Our approach is particularly related to methods that perform 3D object reconstruction by predicting depth images from given viewpoints~\cite{DBLP:conf/eccv/TatarchenkoDB16,DBLP:conf/cvpr/SoltaniH0KT17}. However, these approaches simply fuse the multi-view predictions and fail to enforce the multi-view consistency. Therefore, the reconstruction is blurry and unrealistic. 

\noindent\textbf{Consistent Multi-View Generation.}
A challenge in text-to-textured-mesh models is ensuring the multi-view consistency among the text-to-image outputs. Dreamfusion~\cite{dreamfusion} optimizes the geometry and color of a NeRF through SDS loss.  
Zero123~\cite{zero123} successfully employs a camera transformation as condition to synthesize novel views from an image. TextMesh~\cite{TextMesh} leverages quartet canonical views for mesh rendering, yet it tent to yield blurry results. Several methods ~\cite{Consistent123,Consistent-1-to-3,Mvdream,SyncDreamer,dmv3d,zero123pp,zuo2024videomv} are trained on 3D rendered datasets~\cite{Objaverse} to enhance consistency but tend to lose realistic  details in generated images. Moreover, most of them are agnostic to depth condition, which is crucial for alignment between 2D images and 3D meshes.

\noindent\textbf{Text-to-Texture Models.}
Several pioneers ~\cite{Text2Mesh,CLIP_mesh,TANGO} are trained on a semantic CLIP loss, but still struggle to generate high quality textures with text condition. Trained with a large dataset~\cite{LAION-5B}, diffusion models~\cite{sd} can generate high-quality images with rich details from text prompts. On this basis, additional conditions are added to provide 3D structure-aware through rendering of meshes, such as depth maps and inpainting masks~\cite{control-net}.
Some methods~\cite{chen2023text2tex,10.1145/3588432.3591503} generate single view 2D images on these conditions and incrementally merge them into one mesh texture. However, these approaches can easily lead to suboptimal results due to accumulated errors and the lack of global perception. Using a latent UV map~\cite{latent-nerf,Cao_2023_ICCV,consistent-latent-diff,decorate3d,syncmvd} as a unified representation has demonstrated its effectiveness of increasing global consistency. Since latent pixels compress a patch in RGB images, the decoded RGB pixels cannot guarantee 3D consistency, which leads to the blurriness or inconsistent details in the textures. The other methods~\cite{Yu_2023_ICCV,zeng2023paint3d} finetune a diffusion model on latent UV space to avoid projection, but also generate artifacts with poor UV expansions. Besides, generating PRB materials is also considered as a focus in some work~\cite{deng2024flashtex,youwang2023paintit,Chen_2023_ICCV_Fantasia3D}.
In contrast, our approach performs alternating optimization of non-rigid alignment among the selected images, as well as joint optimization of the best cuts to guide the non-rigid alignment. 

\noindent\textbf{View Selection.}
The quality of a multi-view representation is highly dependent on the distribution of source 2D views. View selection is related to the view saliency problem, which has been extensively studied in the existing literature~\cite{10.1145/1186822.1073244,10.1145/1877808.1877819,DBLP:journals/cgf/ChristieON08,10.1145/2019627.2019628,DBLP:journals/pami/LeifmanST16,DBLP:journals/cgf/KimTLPK17,DBLP:journals/ijcv/SongZZL22}. Closely related to our work, Sun~\etal~\cite{DBLP:conf/cvpr/SunHHG021} present an approach to optimize camera poses to maximize the coverage of an input scene. Wei~\etal~\cite{DBLP:conf/cvpr/WeiYS20} optimize viewpoints to maximize the quality of fusion of the learned 2D features. We introduce a view selection approach that selects a subset of consistent views. Driven by the goal of enforcing multi-view consistency, we propose a novel formulation that optimizes a view-consistent score among the chosen images, while constraining that the selected images provide complete coverage of the underlying 3D model. 

\noindent\textbf{Texture Cutting and Stitching.}
Cutting and stitching overlapping images is a fundamental approach to texture synthesis. The pioneering work of Efros and Freeman~\cite{10.1145/3596711.3596771} optimizes cuts between pairs of overlapping patches for texture synthesis. Our method formulates cut optimization and texture stitching as a second-order MRF problem on the mesh faces, where the second-order term models view consistency across different images. This MRF formulation is introduced by Waechter~\etal~\cite{DBLP:conf/eccv/WaechterMG14}, and our contribution lies in the jointly aligned images as input. In particular, we find it essential to alternate between solving the MRF problem and using the MRF output to refine its inputs. 

%% file: 03_overview.tex
\section{Overview}

\subsection{Problem Statement}

Suppose that we have a 3D model represented as a triangular mesh $\set{M} = (\set{V},\set{F})$ where $\set{V}$ and $\set{F}$ denote the vertex and face sets, respectively. Given a text prompt as input, our goal is to convert $\set{M}$ into a textured mesh. 

\subsection{Approach Overview}

\textbf{View generation.} The first stage takes an overcomplete set of $n$ views of the 3D model $\set{M}$ and the associated depth images as input and outputs the corresponding RGB-D images using a text-to-image model. Consistent with TexFusion~\cite{Cao_2023_ICCV}, we exploit a depth conditioning image generation model. At this stage, the output RGB-D images are generated in isolation, and inconsistencies across views are expected. 

\noindent\textbf{View selection.} The second stage selects a subset of RGB-D images that are consistent with each other. This is formulated as a solution to a quadratic assignment problem in which the objective function models the consistency scores between pairs of RGB-D images, under the constraint that the selected images provide a complete coverage of $\set{M}$. We introduce a novel sequential semidefinite programming approach to solve this problem accurately and efficiently.  

\noindent\textbf{View alignment.} The third stage performs joint non-rigid warping of the selected images so that the pixel colors after warping agree with each other. We first solve a joint optimization problem to alleviate illumination changes in the selected images. We then use SIFTFlow~\cite{DBLP:journals/pami/LiuYT11} to precompute dense correspondences between overlapping views. The resulting correspondences are fed into another joint optimization problem to optimize the deformation of each view so that the deformed images are view-consistent. 

\noindent\textbf{View cutting and stitching.} The final stage outputs a cut of each aligned image so that they form a seamless textured output. We formulate this as solving a second-order MRF problem to optimize the label of each mesh face, where the pairwise terms model view consistency.  The output is a textured mesh with aligned images as the texture source. 

The output provides cuts between pairs of overlapping images along which the pixel colors of the deformed images agree. We then use these cuts to guide the view alignment to focus on regions near these cuts to make the input to the MRF more consistent. Empirically, we find 2 to 3 iterations of view alignment and view cutting and stitching sufficient. 

\begin{figure*}[t]
\includegraphics[width=1.0\textwidth]{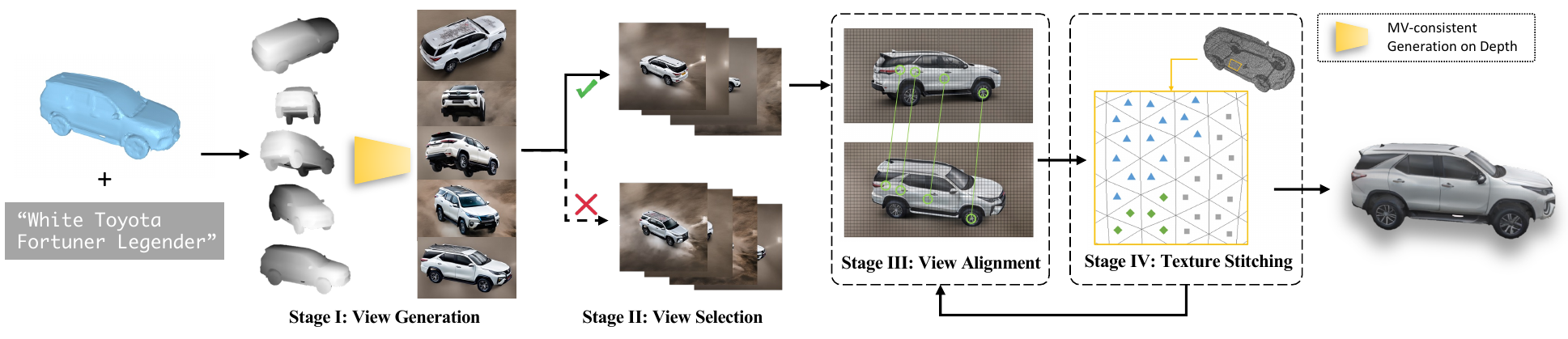}
\caption{The overall pipeline of our approach, which consists of four stages. The first stage (view generation) uses a pre-trained text-to-image model to generate an over-complete set of RGB-D images of the input model. The second stage (view selection) selects a subset of consistent RGB-D images from the view generation output. The third stage (view alignment) performs non-rigid warping to further improve the multi-view consistency among the selected images. The last stage (texture stitching) produces cuts between pairs of overlapping images by solving a second-order MRF problem.}
\label{Figure:Overview}    
\end{figure*}

%% file: 04_approach.tex
\section{Approach}
\label{Section:Approach}
As illustrated in \cref{Figure:Overview}, our approach consists of four stages.
We first describe view generation and view selection in \cref{Subsec:Stage:I:View:Generation} and \cref{Subsec:Stage:II:View:Selection}, respectively. We then elaborate on view alignment in \cref{Subsec:Stage:III:View:Alignment}. Finally, we introduce texture cutting and stitching and discuss how to alternate between this stage and view alignment in \cref{Subsec:Stage:IV:Texture:Stitching}.

\input{04_1_stage_I}
\input{04_2_stage_II}

\input{04_3_stage_III}

\input{04_4_stage_IV}

%% file: 04_1_stage_I.tex
\subsection{Stage I: View Generation}
\label{Subsec:Stage:I:View:Generation}

The purpose of this stage is to generate initial multi-view images from the input text prompt. Directly generating high-quality images that are also consistent using state-of-the-art diffusion-based techniques is extremely difficult. Existing approaches typically explore a trade-off between single-image quality and multi-view consistency. The difference in our approach is that we have a post-processing step that combines view selection, view alignment, and view stitching. This offers us freedom to design the view generation procedure to maximize the performance of the final output. 

Specifically, our post-processing step allows drifted image details as it performs alignment. However, it cannot create image details. Therefore, the multi-view outputs should contain texture details of the final output. Moreover, view alignment is most effective when the absolute displacements are local, meaning some global consistency among the output multi-views is also important. To this end, we introduce a three-phase diffusion procedure to generate the multi-view outputs. 

Similar to Zero123++~\cite{zero123pp}, the first phase leverages 3D priors to guide the diffusion procedure. We first utilize a multi-view consistent model (see supp. material) finetuned on the GObjaverse~\cite{richdreamer} dataset to provide 3D prior. Note that this model is applied only at the initial diffusion step 800. It ensures the global consistency of the image outputs, yet it has minimal influence on later diffusion steps that are responsible for recovering image details.

\begin{figure*}[t]
\centering
\setlength{\tabcolsep}{2.5pt}
\begin{tabular}{ccc}
\includegraphics[height=0.22\textwidth]{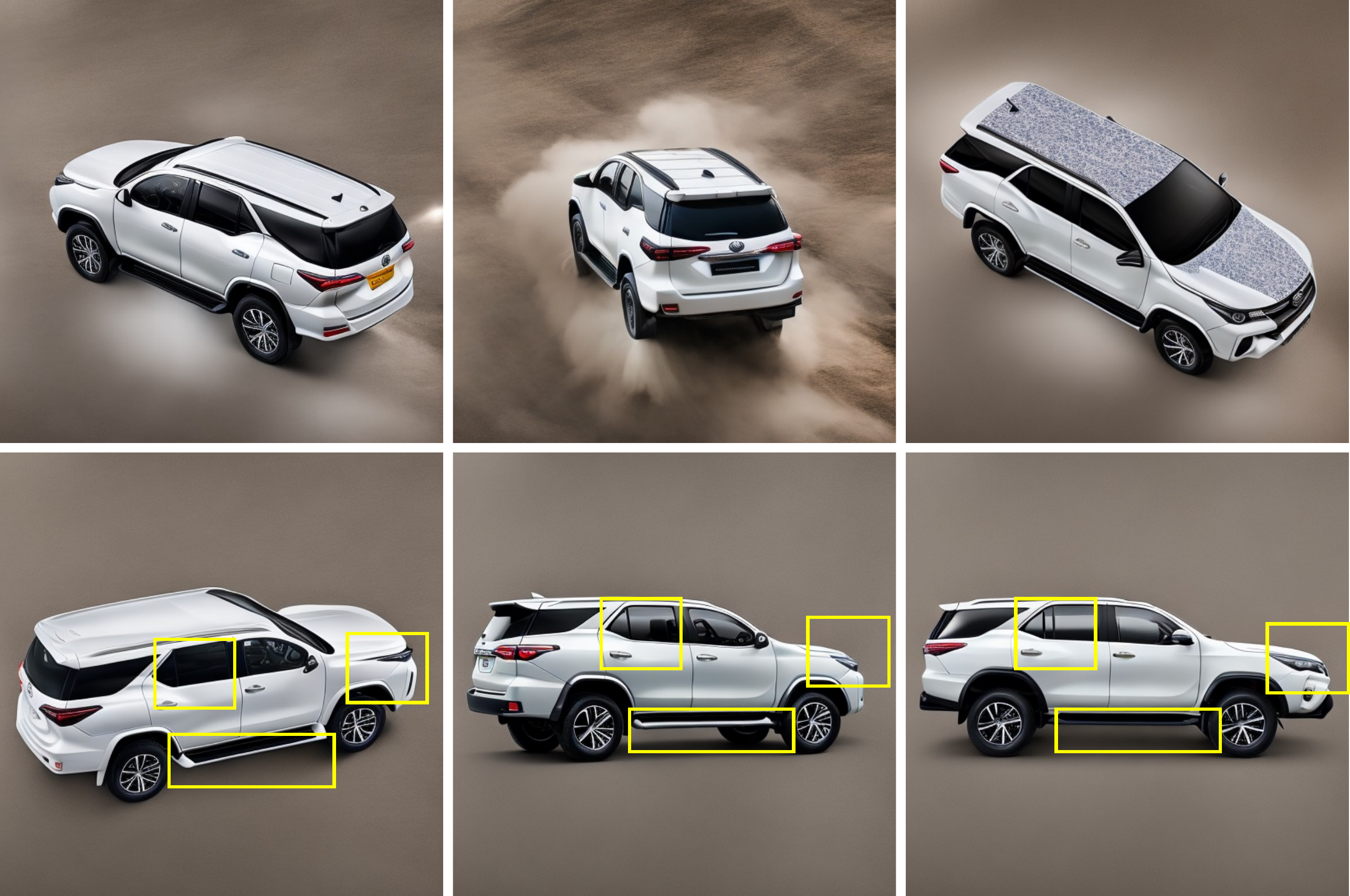}
&\includegraphics[height=0.22\textwidth]{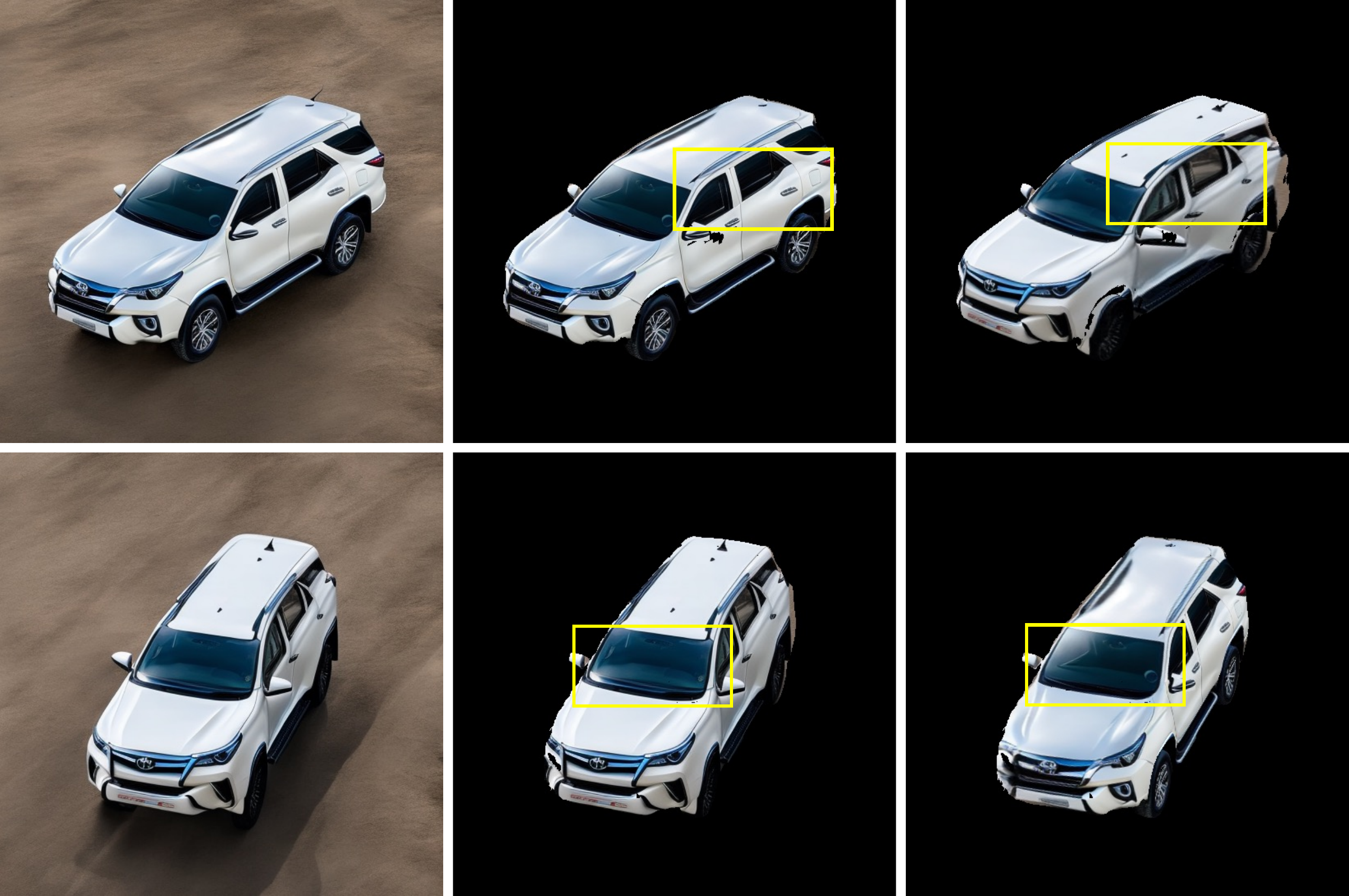}
&\includegraphics[height=0.22\textwidth]{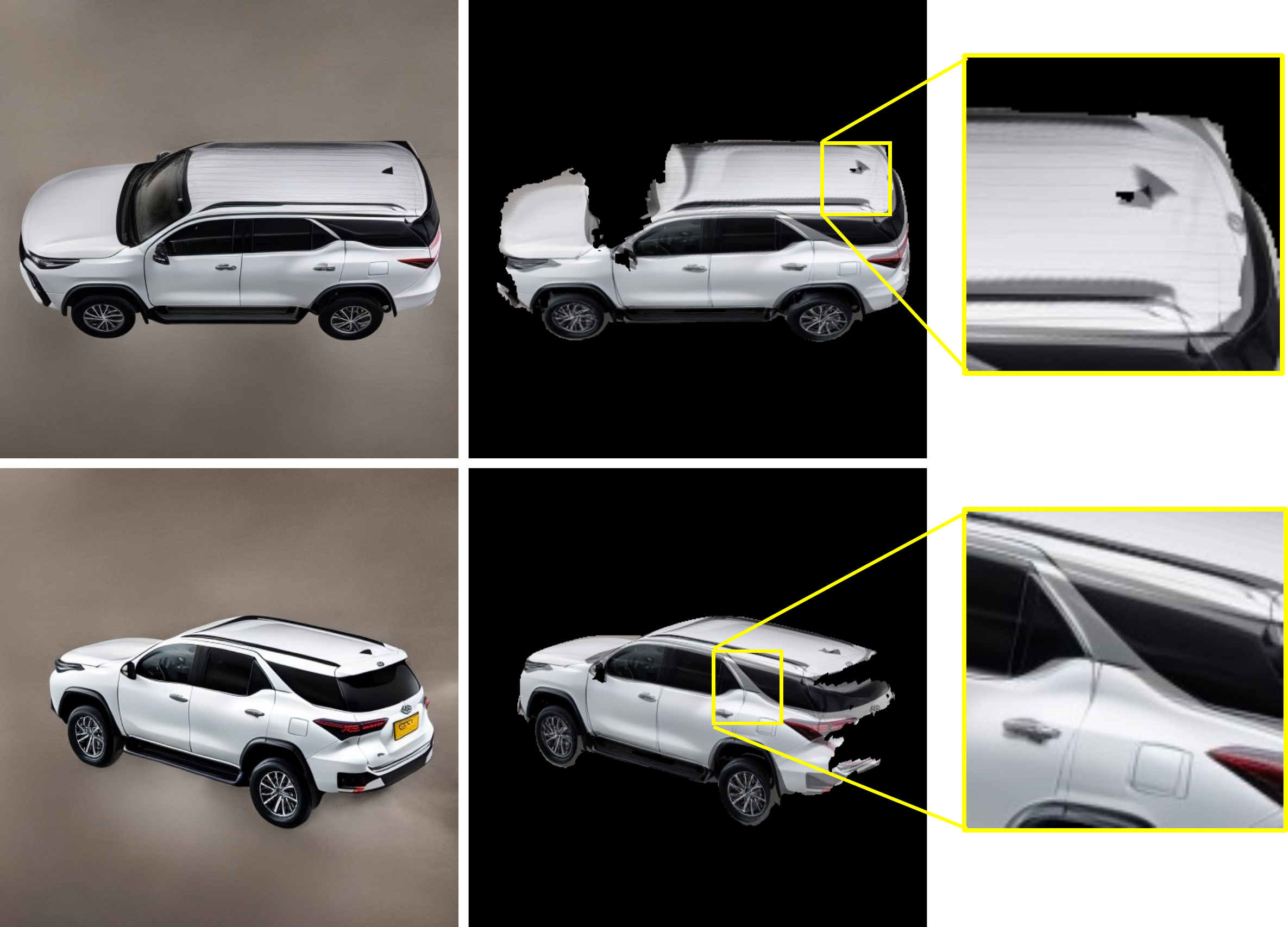}   \\
(a) & (b) & (c) 
\end{tabular}
\caption{\small{Three types of multi-view inconsistencies from pre-trained text-to-image models. (a) The appearance of the image content may change significantly, even with slight perturbations in the camera pose (see the bottom row). (b) There are variations in illumination among overlapping regions of different views. (c) There are drifts in the detail of the image between overlapping regions of different views. (The left column images in (b) and (c) are warped with each other.)}}
\label{Figure:View:Generation}
\end{figure*}

The second phase, which runs between diffusion steps 800-200, adopts latent UV mapping~\cite{latent-nerf}, a popular data-free approach to enforce multi-view consistency. The key idea is to aggregate multiple views in a shared latent map that is then mapped back to individual views. Although improved consistency is achieved, image details can still be dropped during the aggregation step, which performs some type of averaging (see supp. material). Therefore, similar to SyncMVD~\cite{syncmvd}, we do not perform latent UV mapping to generate the final output. 

The final phase performs diffusion on each view independently to generate texture details. 

As shown in \cref{Figure:View:Generation}, the multi-view image generation results are consistent in global style and shape. However, local shapes and styles may differ and there are also drifts. Another important idea is to generate an over-complete set of images and operate on a subset of more consistent images to generate the output. This approach nicely mitigates inconsistency introduced in the third phase, which is performed on each view independently.  

%% file: 04_2_stage_II.tex
\subsection{Stage II: View Selection}
\label{Subsec:Stage:II:View:Selection}

With $\set{I} = \{I_1,\cdots, I_{n}\}$, we denote the result of stage I. The goal of the second stage is to select a subset of images $\set{I}^{\star}\subset \set{I}$ such that 1) the quality of each image is high, 2) the images are mutually consistent, and 3) the images cover the underlying 3D object. In the following, we first introduce a general optimization formulation that takes as input a single image scoring function $s(I_i)$ and a consistency scoring function $s(I_i, I_j)$. We then discuss how to define $s(I_i)$ and $s(I_i, I_j)$ in our setting. 

\noindent\textbf{Optimization formulation.} Given $s(I_i)$ and $s(I_i, I_j)$, we solve a constrained optimization problem to select consistent images. We introduce an indicator variable $z_i\in \{0,1\}$ for each $I_i$, where $z_i = 1$ indicates that $I_i$ is selected and vice versa. Let vector $\bs{z}$ collect all indicators, while vector $\bs{s} = (s(I_i))\in \R^n$ and matrix $S = (s(I_i,I_j))\in \R^{n\times n}$ collect image quality and consistency scores. 

Our goal is to optimize $\bs{z}$ to minimize $\bs{s}^T\bs{z} + \lambda \bs{z}^T S\bs{z}$ while ensuring that the selected images cover the underlying model. To this end, we model the coverage constraint as $C\bs{z}\geq \bs{1}$, that is, each row of $C$ constrains that one face on the underlying mesh is covered by at least one selected image. This leads to the following optimization problem:
\begin{equation}
\underset{\bs{z}\in \{0,1\}^n}{\textup{minimize}} \quad  \bs{s}^T\bs{z} + \lambda \bs{z}^T S\bs{z} \qquad \textup{subject to}  \quad C \bs{z} \geq \bs{1}.
\label{Eq:View:Select:Form}
\end{equation}

\Cref{Eq:View:Select:Form} is hard to optimize, due to non-convexity and integer variables. We introduce two relaxations. The first relaxes $z_i$ so that they can take real values between $0$ and $1$, and the second uses the semidefinite programming relaxation to $\bs{z}^T S\bs{z}$. More precisely, we introduce an additional matrix variable $Y = \bs{z}\bs{z}^T$ and apply the standard relaxation approach as $Y \succeq \bs{z}\bs{z}^T$ and $\textup{diag}(Y) = \bs{z}$. Furthermore, we maintain an index set $\set{I}_f$ that specifies all currently selected images. Initially, we $\set{I}_f = \emptyset$. This leads to the following semidefinite program:
\begin{align}
\underset{\bs{z}\in \R^n, Y\in \R^{n\times n}}{\textup{minimize}} &\quad  \langle \bs{s}, \bs{z} \rangle +  \lambda \langle S, Y\rangle  \nonumber \\
\textup{subject to} & \quad C \bs{z} \geq \bs{1}, \quad 0\leq \bs{z} \leq 1, \quad 0 \leq Y \nonumber,  \quad \textup{diag}(Y) = \bs{z}, \\ 
& \quad z_i = 1, \quad \forall i \in \set{I}_f,
\quad \left(
\begin{array}{cc}
Y & \bs{z} \\
\bs{z}^T & 1
\end{array}
\right) \succeq 0 \label{Eq:SDP:Cons},
\end{align}
where $Y\succeq \bs{z}\bs{z}^T$ is identical to \cref{Eq:SDP:Cons}. 
We then augment $\set{I}_f$ with $i^{\star} = \textup{argmax}_{i\notin \set{I}_f} z_i$. This procedure is iterated until $C\bs{z}\geq 1$ is satisfied.  For simplicity, we assume that the selected views are the first $k < n$ images $I_i,1\leq i \leq k$.

\begin{figure}[t]
\includegraphics[width=0.45\columnwidth]{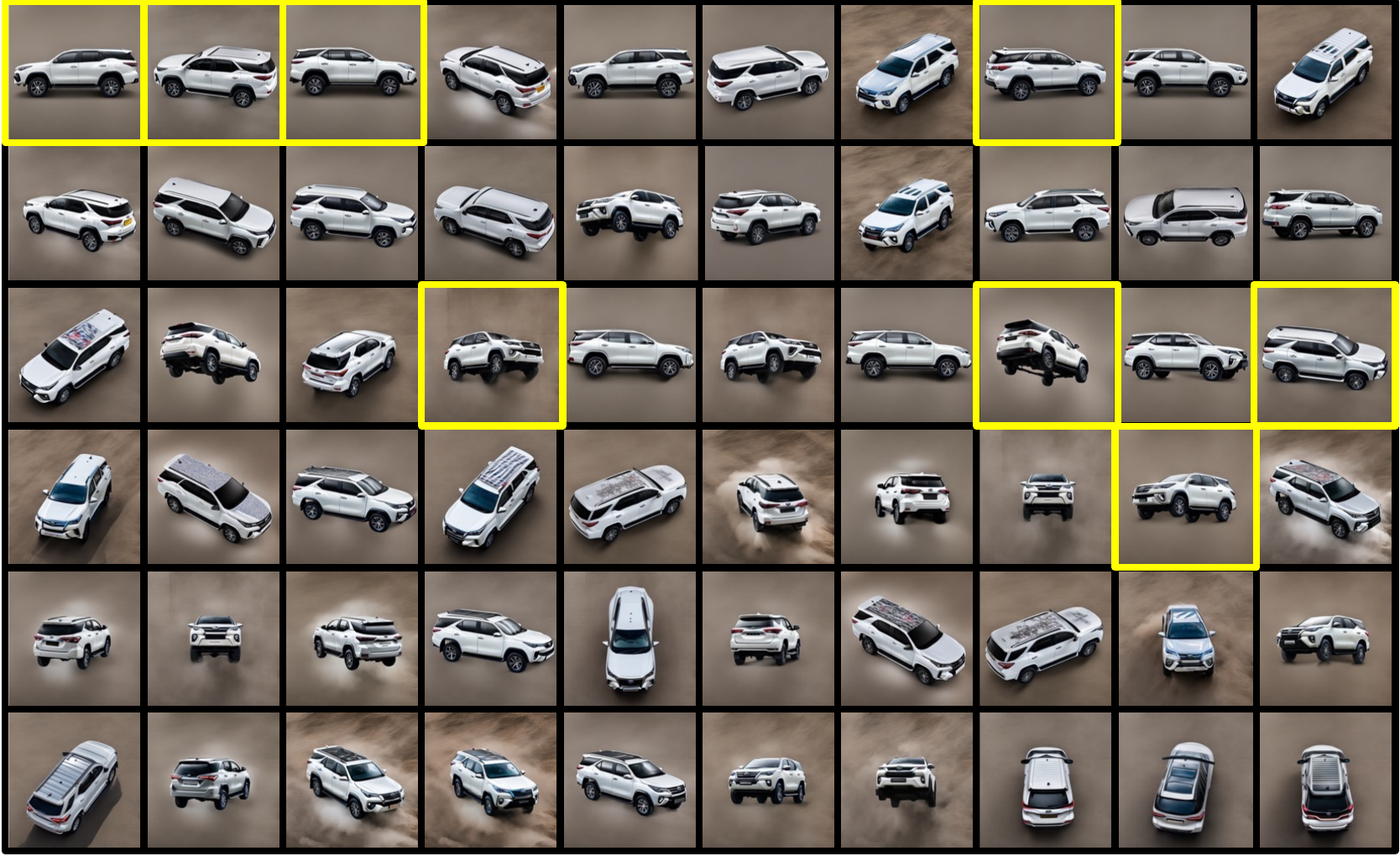}
\includegraphics[width=0.55\columnwidth]{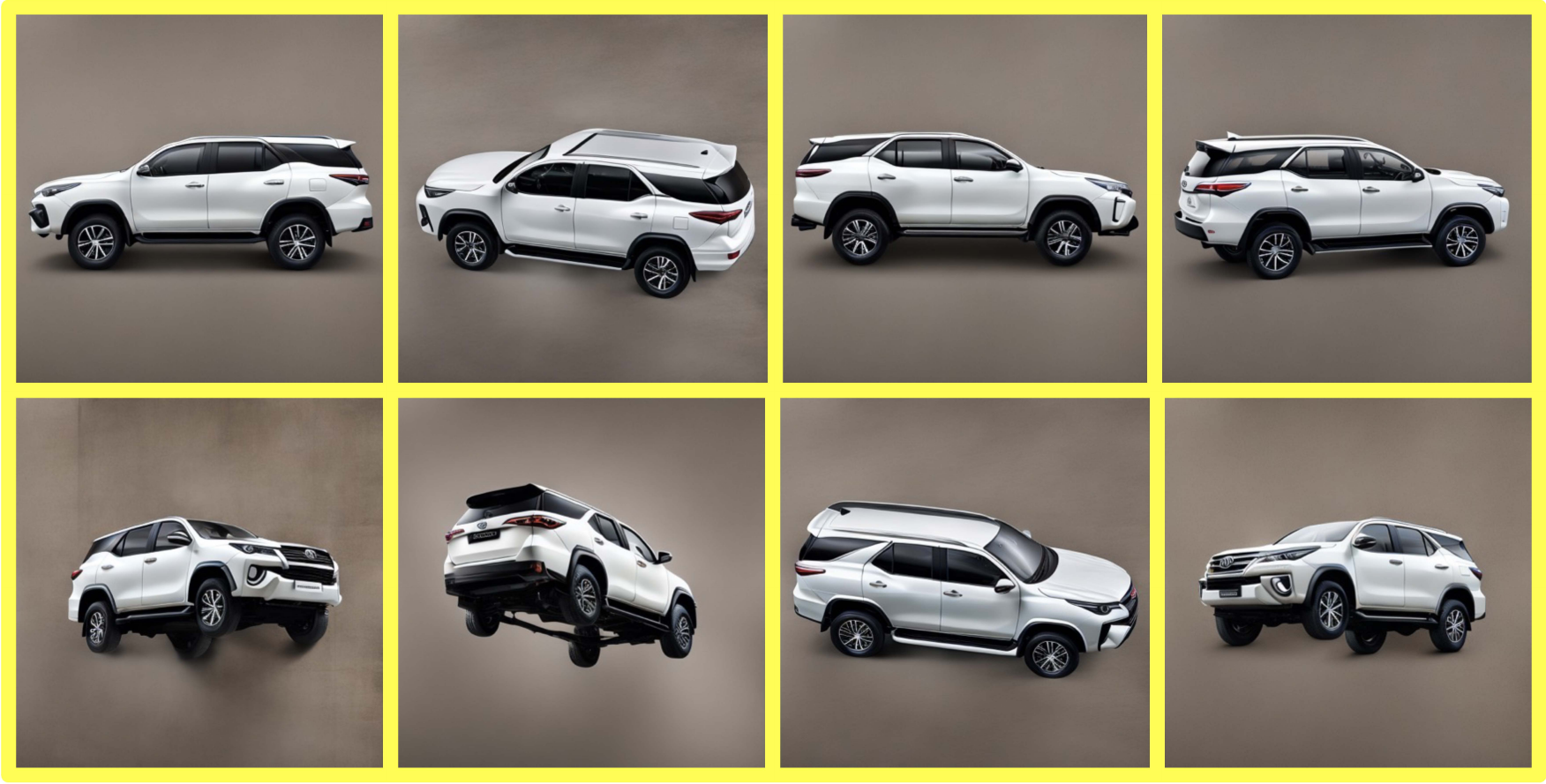}
\caption{\small{Illustration of view selection. (Left) The input images ranked in the increasing order of $S(I_i)$. (Right) The selected images that consider the image scores $S(I_i)$, the consistency scores $S(I_i, I_j)$ and the covering constraint.}}
\label{Figure:View:Selection}
\end{figure}

\noindent\textbf{Image score $s(I)$.} Our key observation is that if the quality of an image $I$ is high, then the pixel colors tend to repeat among other input images. This motivates us to build a probabilistic color model for each vertex $v\in \set{V}$ (\cf, \cref{Figure:View:Selection}~(Left)). Specifically, denote $\set{I}_{v}\subset \set{I}$ as the subset of images in which $v$ is visible. Let $\bs{c}(v,I)$ be the color of $v$ in $I$. We first perform mean shift clustering~\cite{DBLP:journals/pami/ComaniciuM02} to extract the largest cluster of $\{\bs{c}(v,I)|I\in \set{I}_v\}$, which represents the most represented color. With $\bs{c}_v$ and $\sigma_v$ we denote the center and variance of this cluster. We then define
\begin{equation}
p(v,\bs{c}) = \frac{1}{(2\pi)^{\frac{3}{2}}}\frac{1}{\sigma_{v}^3}\exp(-\frac{\|\bs{c}-\bs{c}_{v}\|^2}{2\sigma_{v}^2}).    
\label{Eq:GMM:Model}
\end{equation}
Given an image $I$, let $\set{V}_{I}\subset \set{V}$ collect all visible vertices in $I$. We sort the values $\{p(v,\bs{c}(v, I))|v\in \set{V}_I\}$ in increasing order and set the score $s'(I)$ of $I$ as the $t\%$ percentile ($t=20$) of $p(v,\bs{c}(v, I))$. The final score is given by $s(I) = (s_{\max}-s'(I))/(s_{\max}-s_{\min})$ where $s_{\max}(s_{\min}) = \max_{I}(\min_{I}) s'(I)$. As shown in \cref{Figure:View:Selection}~(Left), high-quality images tend to have small values in $s(I)$.

\noindent\textbf{Consistency score $s(I_i, I_j)$.}  
For the purpose of view selection, we find that defining $s(I_i, I_j)$ based on mean color differences without alignments is sufficient.

We set $s(I_i, I_j) = 0$ if $I_i$ and $I_j$ do not overlap. \Cref{Figure:View:Selection}~(Right) shows that the consistency score is critical to selecting consistent images.

%% file: 04_3_stage_III.tex
\subsection{Stage III: View Alignment}
\label{Subsec:Stage:III:View:Alignment}

\begin{figure*}[t]
\centering
\setlength{\tabcolsep}{2.5pt}
\begin{tabular}{cccc}
\includegraphics[height=0.275\textwidth]{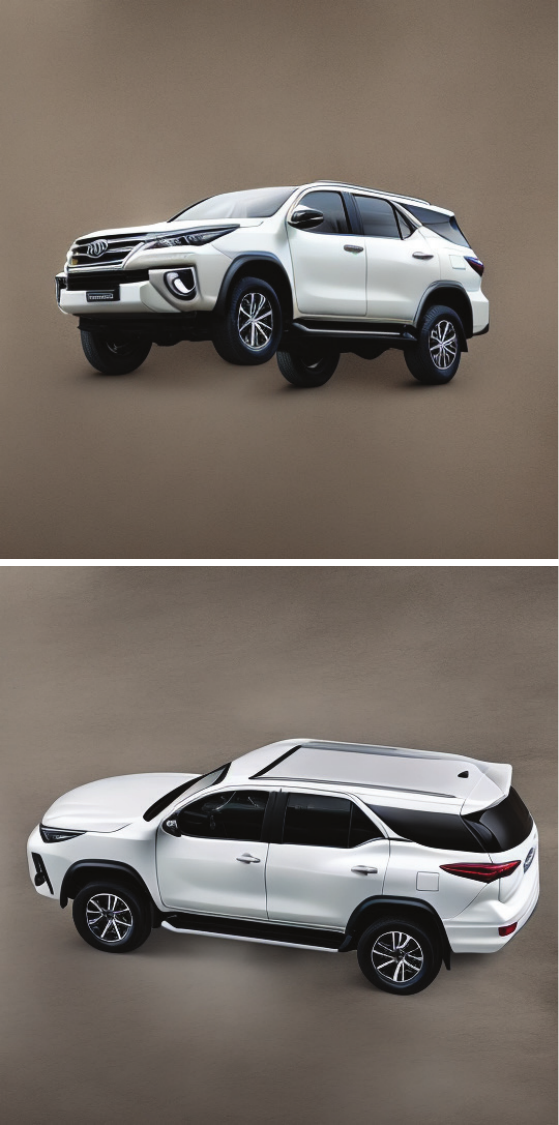} &
\includegraphics[height=0.275\textwidth]{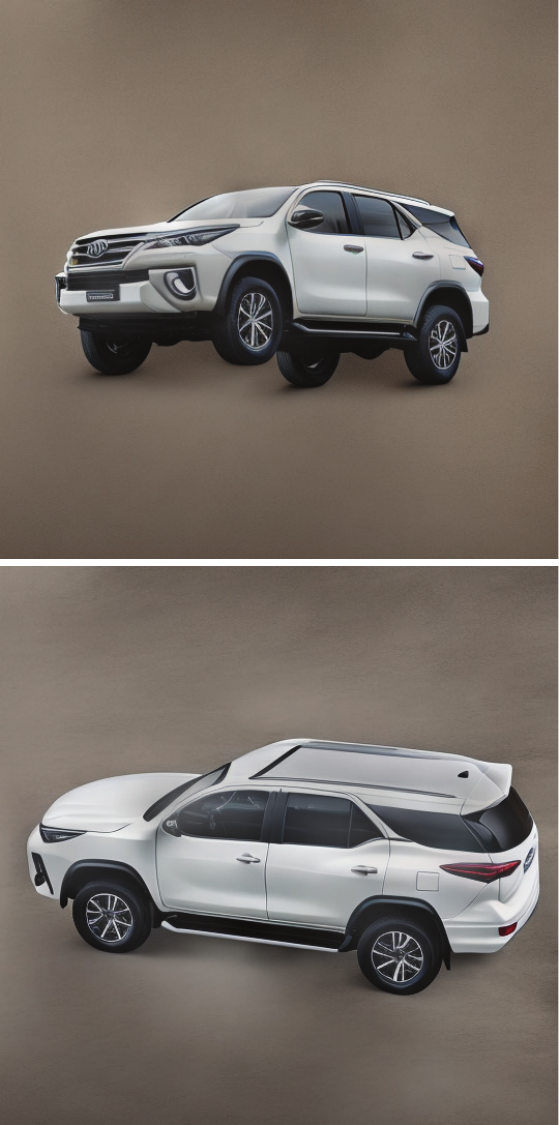} &
\includegraphics[height=0.275\textwidth]{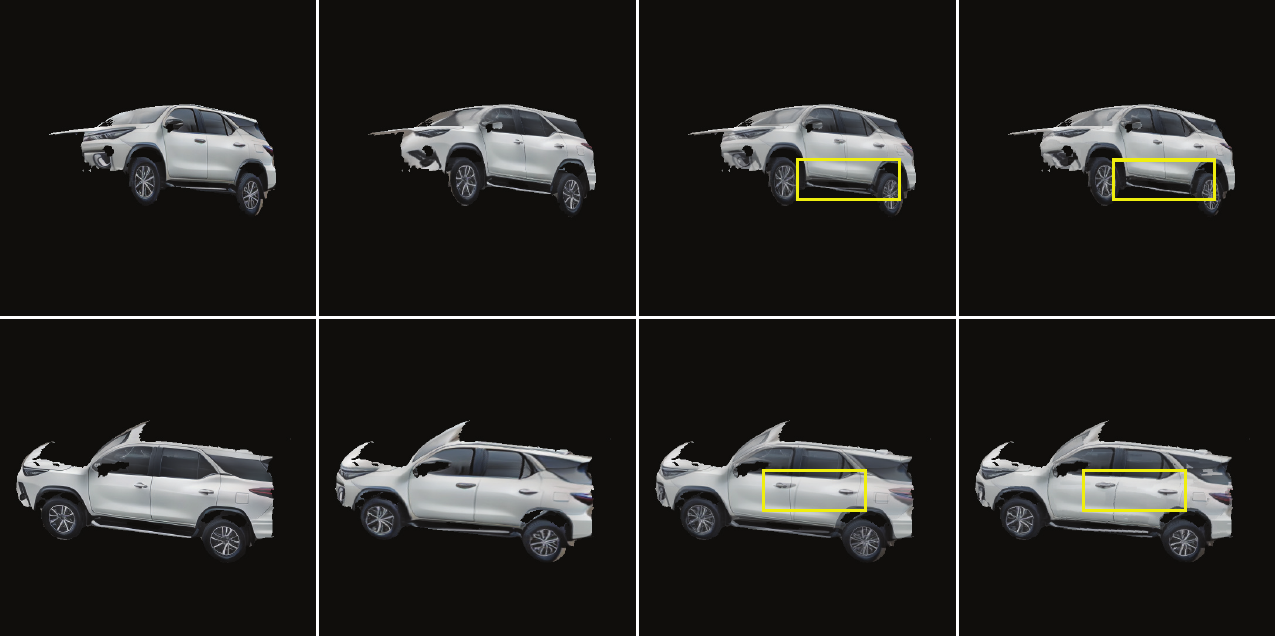} &
\includegraphics[height=0.275\textwidth]{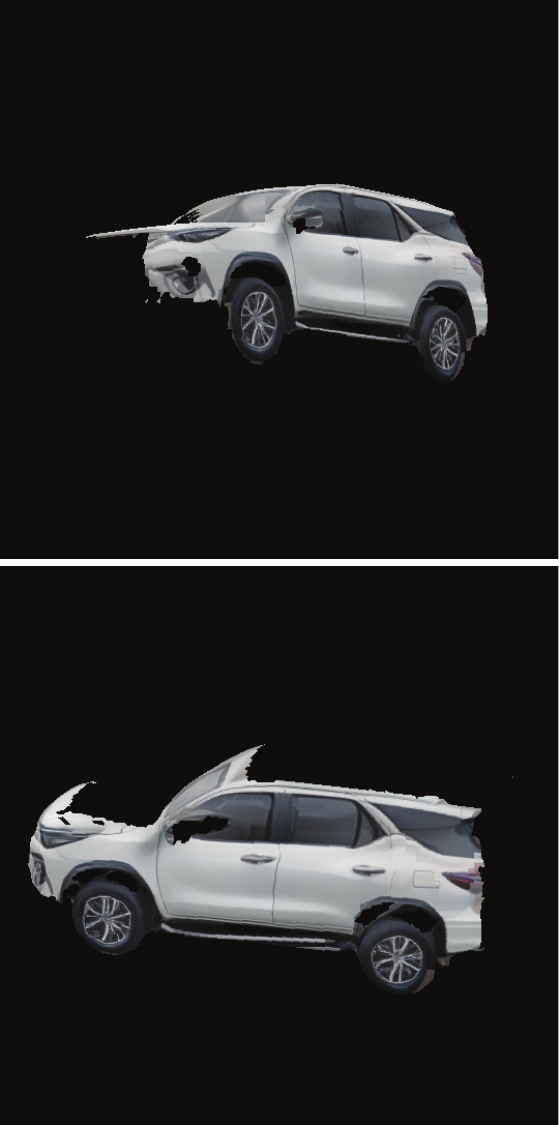} \\
(a)     & (b)  & (c) & (d)
\end{tabular}
\caption{\small{Illustration of the view alignment stage. We show the result of two overlapping images through the joint view alignment procedure. (a) Two input images. (b) After adjustment of the global illumination. (c) From left to right: overlapping regions of each input image, warped images using ground-truth maps between the overlapping regions, overlaid results of the first two columns, and overlaid after SIFTFlow alignment. (d) Overlaid alignments after joint alignment. The pairwise SIFTFlow alignments are preserved.}}
\label{fig:View:Alignment}
\end{figure*}

The third stage of our approach performs non-rigid alignment among the selected RGB-D images, making them consistent among their overlapping regions. As illustrated by \cref{fig:View:Alignment}, we achieve this goal in two steps. The first step adjusts the pixel colors together to reduce the illumination variations in the selected images. The second step performs joint image warping to align image features across the selected images. In both steps, we parameterize the deformation of each image using the free-form deformation (FFD)~\cite{10.1145/15922.15903} on a regular $L\times L$ lattice (we choose $L = 20$ for color adjustment and $L = 40$ for image warping). Note that it is possible to avoid illumination variations by training text-to-image models from rendered 3D shapes. However, our goal is to use photorealistic images from pre-trained text-to-image models. Therefore, we chose to solve an optimization problem to consistently reduce color variations. 

\noindent\textbf{Color adjustment.} With $I_i^x \in \R^{m\times m}$, where $x\in \{r,g,b\}$ denote the color channel the image $I_i$, we parameterize the adjusted image $\hat{I}_i^x$ as:
\begin{equation}
\hat{I}_i^x(\bs{p}) = B(\bs{p})\bs{c}_{i, x}^{1} + I_i^x(\bs{p})B(\bs{p})\big(\bs{1}-\bs{c}_{i, x}^{0}-\bs{c}_{i, x}^{1}\big),
\end{equation}
where $B(\bs{p})\in \R^{1\times L^2}$ is a sparse vector that encodes the FFD coefficients of pixel $\bs{p}$; $\bs{c}_{i, x}^{0}$ and $\bs{c}_{i, x}^{1}$ are control weights that correspond to all black and white pixels. Let $M_i$ be the diagonal matrix that specifies the inactive control weights (\ie, those that correspond to the background pixels). The control weights are restricted by $0\leq \bs{c}_{i, x}^{0}$, $0\leq \bs{c}_{i, x}^{1}$, $\bs{c}_{i, x}^{0}+\bs{c}_{i, x}^{1}\leq \bs{1}$, and $M_i\bs{c}_{i, x}^{0} = M_i\bs{c}_{i, x}^{1} = \bs{0}$, which ensure that $\hat{I}_i^x\in [0,1]^{m\times m}$.

We use $(I_{ij}\subset I_i, I_{ji}\subset I_j)$ to denote the regions that overlap between $I_i$ and $I_j$. Let $g_{ij}: I_{ji}\rightarrow I_{ij}$ and $g_{ji}: I_{ij}\rightarrow I_{ji}$ be the ground-truth correspondence maps between them. We solve the following quadratic program with linear constraints to determine $\bs{c}_{i, x}^{0}$ and $\bs{c}_{i, x}^{1}$:
\begin{align}
\min_{\{\bs{c}_{i,x}^{l}\}} & \sum\limits_{x\in \{r,g,b\}}\Big(\sum\limits_{i \neq j} \sum\limits_{\bs{p}\in I_{ij}}\|\hat{I}_i^{x}(\bs{p}) - \hat{I}_{j}^x(g_{ij}(\bs{p}))\|^{2} + \mu\sum\limits_{i=1}^{k}\sum\limits_{l=0}^1 {\bs{c}_{i,x}^{l}}^TL{\bs{c}_{i,x}^{l}} \Big)
\label{Eq:Obj:Func:Color:Adjustment}    \nonumber \\
s.t. & \quad \bs{0}\leq \bs{c}_{i, x}^{0}, \bs{0}\leq \bs{c}_{i, x}^{1}, \bs{c}_{i, x}^{0}+\bs{c}_{i, x}^{1}\leq \bs{1}, \quad M_i\bs{c}_{i, x}^{0} = M_i\bs{c}_{i, x}^{1} = \bs{0},
\end{align}
where $L$ is a Laplacian matrix of the control lattice. \Cref{fig:View:Alignment}~(b) shows that this step can effectively reduce the variation of illumination between overlapping images.  

\noindent\textbf{Image warping.} Let $(O_{ij}\subset I_{ij}, O_{ji}\subset I_{ji})$ be the focused overlapping subsets between $I_i$ and $I_j$. Our goal is to deform each $I_i$ so that $O_{ij}$ and $O_{ji}$ are aligned among the deformed images. Initially, $O_{ij} = I_{ij}$ and $O_{ji} = I_{ji}$. After determining the optimal cut between $I_i$ and $I_j$, we recompute $O_{ij}$ and $O_{ji}$ as the dilated regions along the cuts. This allows us to focus on alignments among potentially more consistent regions to improve alignment results. 

Let $f_{ij}: g_{ij}(O_{ji})\rightarrow O_{ij}$ and $f_{ji}: g_{ji}(O_{ij}) \rightarrow O_{ji}$ be dense pixel-wise alignments between the focused overlapping regions under the view of each image. In our experiments, we compute $f_{ij}$ and $f_{ji}$ using SIFTFlow~\cite{DBLP:journals/pami/LiuYT11}, a state-of-the-art method that establishes high-fidelity dense correspondences (\cref{fig:View:Alignment}~(c)).  

Let $\bs{c}_i \in \R^{2\times L^2}$ be the grid of control points that determine the deformation $\set{D}_i$ of the image $I_i$. The deformed position of $\bs{p}$ is given by $\set{D}_i(\bs{p}) = \bs{c}_i \bs{b}(\bs{p})$. Let $\bs{c}_i^{(0)}$ be the initial control grid of $\set{D}_i$, we model the following optimization problem to optimize $\{\bs{c}_i\}$ jointly:
\begin{align}
\min_{\{\bs{c}_i\}} & \sum\limits_{i\neq j} \sum\limits_{\bs{p}\in O_{ij}}\|\bs{c}_i\bs{b}(\bs{p}) - g_{ij}(\bs{c}_j\bs{b}(g_{ji}(\bs{p}))) -f_{ij}(\bs{p})\|^{2} \nonumber \\
& + \sum\limits_{i=1}^{k}\Big( \mu_1 \|\bs{c}_i-\bs{c}_i^{(0)}\|^2 + \mu_2\textup{Tr}(\bs{c}_iL\bs{c}_i^T) \Big),
\label{Eq:Obj:Func}    
\end{align} 
where $\mu_1 = 2$ and $\mu_2 = 200$ in all our experiments. Since \cref{Eq:Obj:Func} consists of non-linear least squares, we apply the Gauss-Newton method for optimization. 

\Cref{fig:View:Alignment}~(d) shows that joint view alignment offers deformations of each image that preserve the alignments provided by pairwise SIFT flows.

%% file: 04_4_stage_IV.tex
\subsection{Stage IV: Texture Stitching}
\label{Subsec:Stage:IV:Texture:Stitching}

\begin{wrapfigure}{r}{0.5\textwidth}

\scriptsize
\includegraphics[width=0.48\columnwidth]{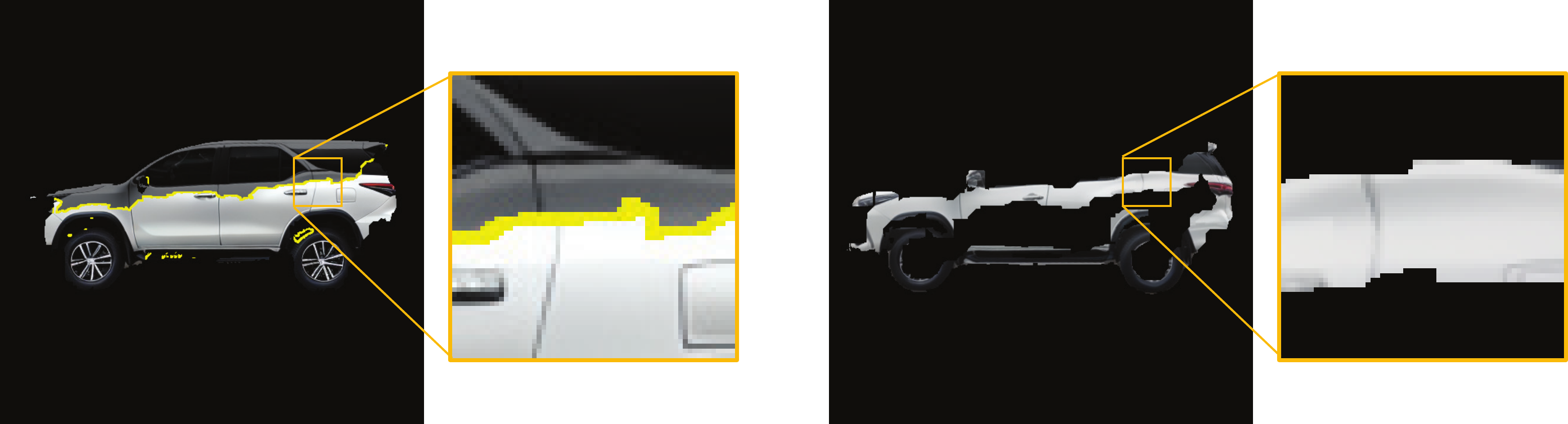}
\caption{\small{(Left) Optimized cut among two overlapping images showing mis-alignment along the seams of different images. (Right) Overlaid SIFTFlow alignment results among the masked regions, where the seams are aligned.}}
\label{Figure:Overlapping:Refinement}    
\end{wrapfigure}

The final stage of our approach takes the aligned RGB-D images $I_i^{\star} = \set{D}_i^{\star}(I_i), 1\leq i \leq k$ as input and outputs a complete textured 3D model. This is formulated as decomposing the mesh faces $\set{F}$ of the input mesh $\set{M}$ into sets $k$: $\set{F} = \set{F}_1\cup \cdots \set{F}_{k}$ where $\set{F}_i$ is associated with $I_i^{\star}$. We formulate this procedure as a joint labeling problem. Specifically, we associate each face $f\in \set{F}$ with a random variable $z_f\in \{1,\cdots, k\}$ that specifies the set of faces $\set{F}_{z_f}$ to which it belongs. Let $\set{E} = \{(f,f')\}$ collect all pairs of adjacent faces. We optimize $z_f$ by minimizing the following objective function:
\begin{equation}
\min\limits_{\{z_f\}} \sum\limits_{f\in \set{F}} \phi_{f}(z_f) + \mu \sum\limits_{(f,f')\in \set{E}} \phi_{ff'}(z_f, z_{f'})    
\label{Eq:Labeling}
\end{equation}
where $\phi_f(z_f)$ is the cost of label $z_f$, and $\phi_{ff'}(z_f, z_{f'})$ is the consistency between $z_f$ and $z_{f'}$. In this paper, we define $\phi_f(z_f)$ using a variant of \cref{Eq:GMM:Model}, that is, a label has a small cost if it is more consistent with other images. Furthermore, $\phi_{ff'}(z_f, z_{f'}) = 0$ if $z_f = z_{f'}$. Otherwise, its value depends on the color difference between $z_f$ and $z_{f'}$. Details are deferred to the supp. material.

We solve \cref{Eq:Labeling} using the OPENGM2~\cite{kappes-2015-ijcv} implementation of TRWS~\cite{DBLP:journals/pami/Kolmogorov06}. \Cref{Figure:Overlapping:Refinement}~(Left) shows an example of optimized cuts between aligned images.

After obtaining $\set{F}_i,1 \leq i \leq k$, we use the corresponding cut between each pair of overlapping images $I_i$ and $I_j$ to recompute $O_{ij}$ and $O_{ji}$. This is achieved by dilating the cut in each $I_i$ by $l$ pixels ($l=20$) and taking the union of the dilated regions to define $O_{ij}$ and $O_{ji}$. After that, we rerun the view alignment to generate the inputs for texture stitching. This alternating procedure is iterated until the final output becomes steady (typically 2-3 iterations). \Cref{Figure:Overlapping:Refinement}~(Right) shows that this alternate procedure can enhance visual consistency along the cuts. 

%% file: 05_results.tex
\section{Experimental Results}

\begin{figure*}[t]
\includegraphics[width=1.0\textwidth]{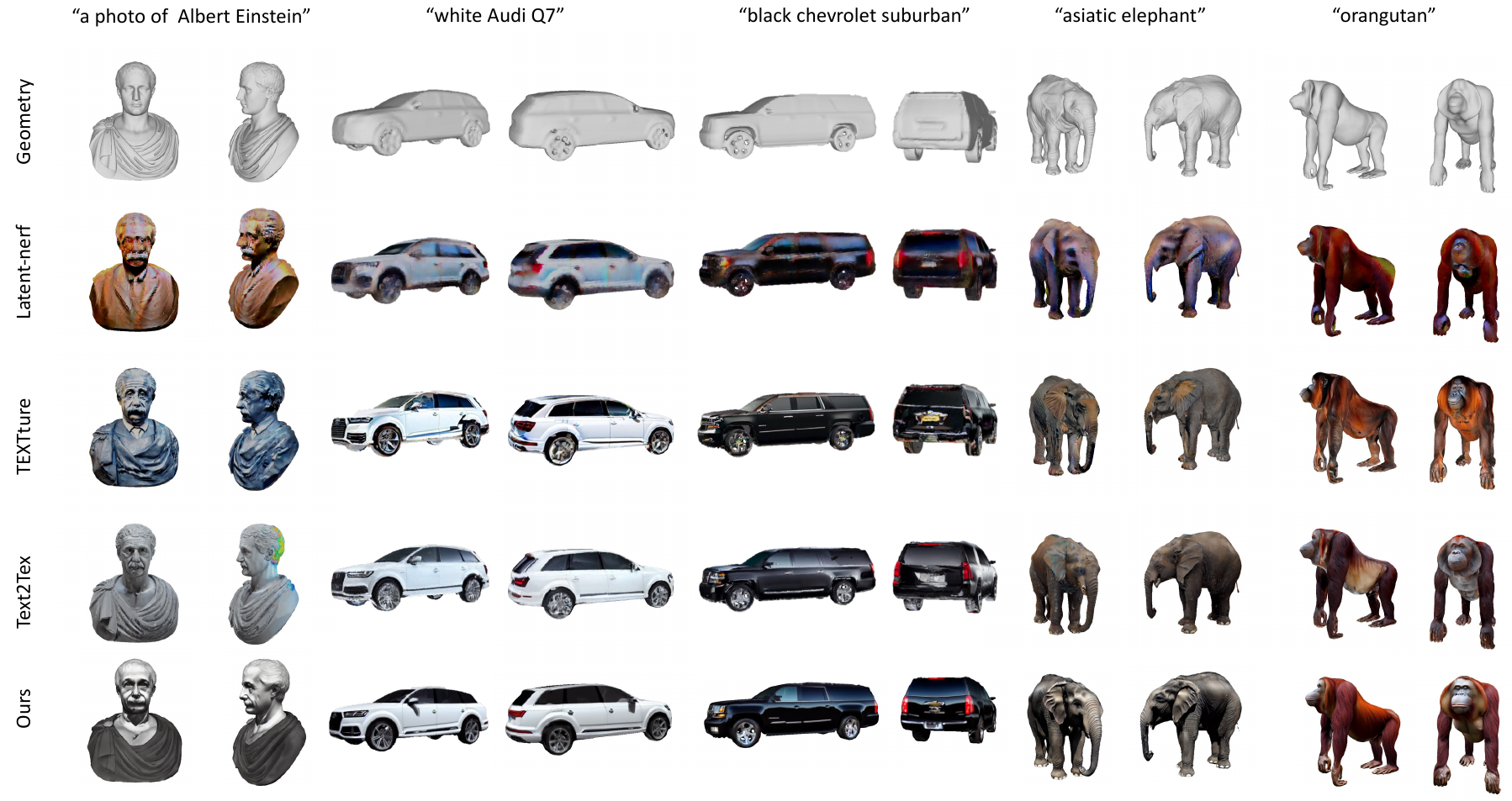}
\caption{\small{Qualitative visualization. From top to bottom: Input 3D mesh; Latent-NeRF; TEXTure; Text2Tex; Ours. }}
\label{Figure:Qualitative:Results}
\end{figure*}

We begin with the experimental setup in \cref{Subsec:Experimental:Setup}.
We then present the experimental results in \cref{Subsec:Analysis:Results}.
Finally, we describe an ablation study in \cref{Subsec:Ablation:Study}.

\subsection{Experimental Setup}
\label{Subsec:Experimental:Setup}

\textbf{Dataset.}
We collect 300 models from different categories in Objaverse~\cite{Objaverse}. We also collect 20 models used in other text-to-texture-mesh methods and models from generation approaches~\cite{richdreamer,Dong_2024_CVPR} for qualitative comparison.

\noindent\textbf{Baseline approaches.} We evaluate our method against Latent-NeRF~\cite{latent-nerf}, TEXTure~\cite{10.1145/3588432.3591503}, Text2Tex~\cite{chen2023text2tex}, SyncMVD~\cite{syncmvd}, and Zero123++~\cite{zero123pp} the state-of-the-art text-to-textured-mesh approaches with open source implementations. We also compare with a popular multi-view texturing method (MVS-Texturing)~\cite{DBLP:conf/eccv/WaechterMG14} to show the superiority of our global optimization framework.

\noindent\textbf{Quantitative Comparisons.}
Qualitative evaluation of generative models is always a difficult problem. We follow TexFusion~\cite{Cao_2023_ICCV} and use the same qualitative evaluation metric (\ie, using the generated image based on the depth condition as an upper bound to calculate the FID~\cite{heusel2017gans}). 
In addition, a user study is designed in which human participants are asked to assign a numerical score between 1 and 10 for the generated result. 

\begin{figure*}[t]
\centering
\setlength{\tabcolsep}{2.5pt}
\includegraphics[width=1.0\textwidth]{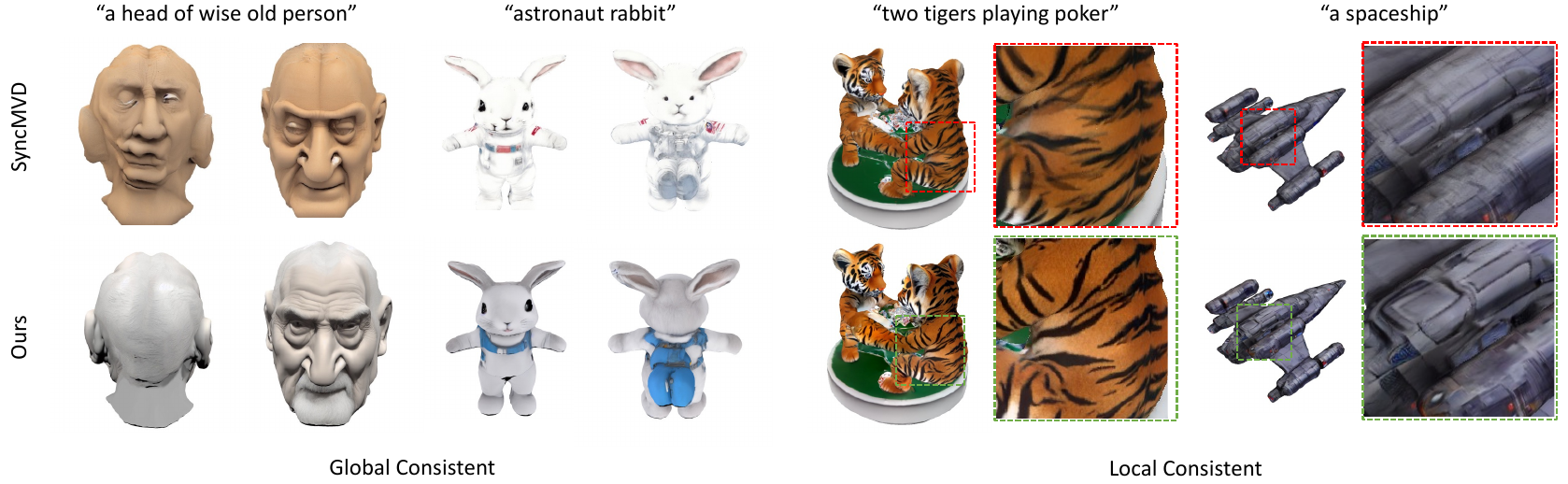}
\caption{\small{Global consistency and local consistency comparisons between SyncMVD and Ours. 
}}
\label{Figure:compare syncMVD}
\end{figure*}

\subsection{Analysis of Results}
\label{Subsec:Analysis:Results}

\begin{wrapfigure}{r}{0.5\textwidth}
\centering
\setlength{\tabcolsep}{2.5pt}
\includegraphics[width=0.48\textwidth]{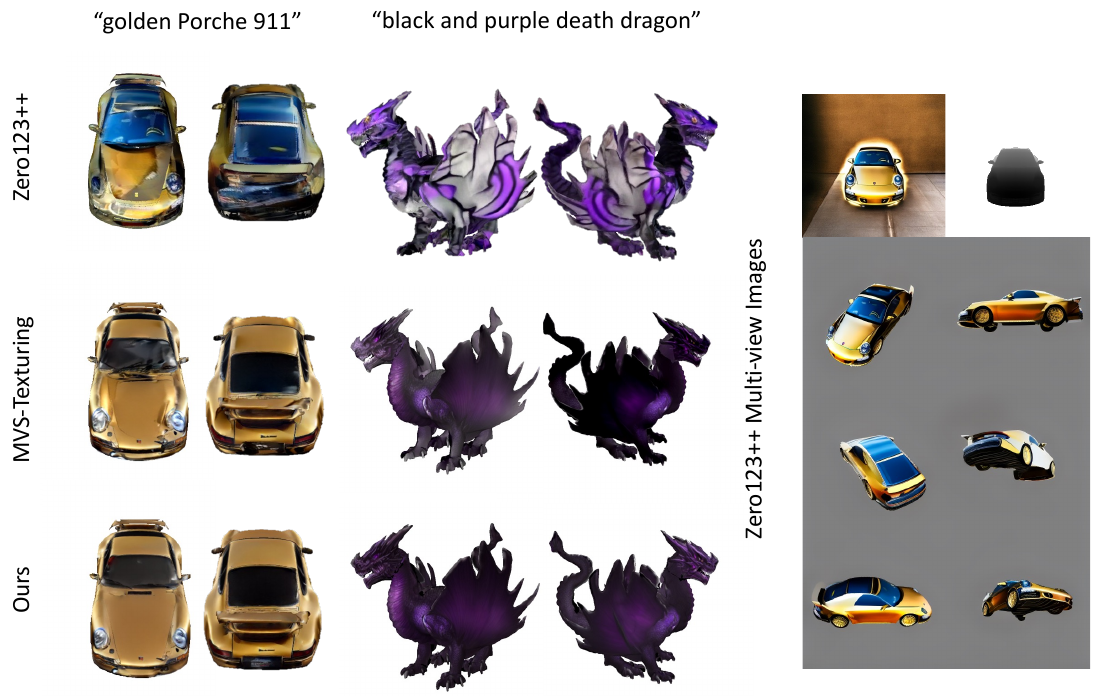}
\caption{\small{The left part: comparison between Zero123++, MVS-Texturing and Ours. The right part: intermediate results of Zero123++.
}}
\label{Figure:compare zero123++}
\end{wrapfigure}

\Cref{Figure:Qualitative:Results} compares our approach to the baseline approaches in five models with text prompts. Visually, our approach preserves the appearance among the individual images while ensuring a globally consistent texture output. In contrast, the results of Latent-NeRF are blurry. This is due to the averaging operation performed on the inconsistent outputs from pre-trained text-to-image models. TEXTure and Text2Tex demonstrate some qualitative improvement over Latent-NeRF. However, the results of TEXTure and Text2Tex still exhibit noticeable visual artifacts due to accumulated errors in the sequential procedure. 

\Cref{Figure:compare syncMVD} compares our approach to SyncMVD~\cite{syncmvd}. Our approach reduce some of the global inconsistencies due to the 3D prior. The details are also clearer thanks to our optimization in rgb space.

\Cref{Figure:compare zero123++} compares our approach to Zero123++~\cite{zero123pp} and MVS-Texturing~\cite{DBLP:conf/eccv/WaechterMG14}. We generate the reference view image for Zero123++ via a control-net~\cite{control-net} and use MVS-Texturing to generate a texture map. It can be observed that our approach is better in detail and global consistency than Zero123++. We compare MVS-Texturing with our optimization under the same dataset generated from stage I. Both MVS-Texturing and our approach solve an MRF problem to obtain a textured output. However, our approach yields more visually appealing results, due to view selection and the alternating optimization between view alignment and view stitching.

\begin{wraptable}{r}{0.7\textwidth}
\caption{FID is computed w.r.t. the set of images synthesized by control-net. User study: numerical scores in four aspects assigned between 1 and 10 (higher is better).}
\resizebox{0.68\textwidth}{!}{
\centering
\begin{tabular}{cccccc}
\hline
\multirow{3}{*}{Method}& \multirow{3}{*}{FID(↓)} & \multicolumn{4}{c}{User study}\\
                        & & \multicolumn{1}{c}{\begin{tabular}[c]{@{}c@{}}Natural \\ Color (↑)\end{tabular}} & \multicolumn{1}{c}{\begin{tabular}[c]{@{}c@{}}More \\ Details (↑)\end{tabular}} & \multicolumn{1}{c}{\begin{tabular}[c]{@{}c@{}}Less \\ Artifacts (↑)\end{tabular}} & \multicolumn{1}{c}{\begin{tabular}[c]{@{}c@{}}Align with \\ Prompt (↑)\end{tabular}} \\ \hline
Latent-NeRF      & 59.71          & 5.43          & 4.38          & 4.84          & 5.56  \\
TEXTure          &  44.70         & 7.29          & \textbf{7.92} & 6.76          & \textbf{8.45}   \\
Text2Tex         &39.35           & 7.56          & 7.33          & 6.91          & 7.68   \\
SynvMVD          &\textbf{\textcolor{blue}{33.92}}           & \textbf{\textcolor{blue}{8.65}}          & 7.21          & \textbf{\textcolor{blue}{7.97}}          & 8.32   \\
Zero123++        &53.93           & 6.80          & 5.83          & 5.44          & 8.09  \\
MVS-Texturing    &37.47           & 7.36          & 6.46          & 5.51          & 8.34   \\
Ours             & \textbf{29.26} & \textbf{8.92} & \textbf{\textcolor{blue}{7.49}}         & \textbf{8.23} & \textbf{\textcolor{blue}{8.35}} \\
\hline
Ablation Study & \multicolumn{1}{c}{\begin{tabular}[c]{@{}c@{}}No view  \\ selection\end{tabular}}  & \multicolumn{1}{c}{\begin{tabular}[c]{@{}c@{}}No view   \\ alignment\end{tabular}}  & \multicolumn{1}{c}{\begin{tabular}[c]{@{}c@{}}No   alternating \\ optimization\end{tabular}} & Ours 
\\
FID(↓) &41.59&35.24&31.55&29.26 \\ \hline
\end{tabular}}
\label{table:user_study}
\end{wraptable}

As shown in \cref{table:user_study}, the quantitative results are consistent with the relative improvement in visual performance. The blue numbers mean the second best performance. This result quantitatively justifies our design that combines view generation, selection, alignment, and view stitching. 

In the user study, 87 participants participated in the comparison according to four metrics, including the naturalness of the coloring (``natural color''), the details of textures (``more details''), the artifacts (``less artifacts''), and the alignment with the input prompt (``align with prompt''), with each participant undertaking 3 tests.

Our approach takes 10 to 15 minutes to texturize each model. On average, stage I takes 2 to 6 minutes, stage II takes 2.3 minutes, stage III takes 5.2 minutes, and stage IV takes 2 minutes. 
A desktop with a CPU of 3.2GHz and NVIDIA 4090 GPU (only for stage I) is used in our experiments. The post-processing stages (stages II-IV) take approximately twice the running time of stage I, which is comparable to baseline approaches. Increasing the efficiency of stages II to IV, possibly using neural networks, GPU-based parallelism, or a combination of the two, is subject to future research. 

\subsection{Ablation Study}
\label{Subsec:Ablation:Study}

This section describes an ablation study of our approach. 
Quantitative results are given in and \cref{table:user_study} and a qualitative result is available in supp. material.

\noindent\textbf{No view selection.} View alignment is a critical component of our approach. Without view selection, the FID score increases by 42\%. Our observation is that without view selection, the textured mesh has appearance artifacts from erroneously selected images. There are also regions where the texture appearance is discontinuous. The latter issue is caused by inconsistent image alignment, where input affects output, despite the fact that robust alignment is used. 

\noindent\textbf{No view alignment.} View alignment is another critical component of our approach. Without view alignment, the FID score increases by 20\%. In this case, there are discontinuities across the textured mesh. These effects are caused by local multiview inconsistencies among the selected views. This issue cannot be easily addressed by solving an MRF problem to identify the cuts across the most consistent regions.

\noindent\textbf{No alternating optimization.} Without alternating optimization, the performance of our approach drops both qualitatively and quantitatively. The FID score increases by 8\%. As shown in \cref{Figure:Overlapping:Refinement}(a), we see some discontinuities across the cut. These are caused by the view alignment, in which pairwise alignments take into account all overlapping regions, even though some regions disagree. With alternating optimization, view alignments tend to be more consistent along the cuts, leading to improved results after cutting.

%% file: 06_limitations.tex
\section{Limitations}

One limitation of our approach is that we solve an optimization problem to adjust the colors of the generated images to make them consistent across different views. This procedure does not understand the latent factors of the illumination changes. In the future, we plan to learn a network to adjust the color of each image conditioned on a latent parameter, where the latent parameters are optimized together to ensure view consistency among the adjusted images.  

Another limitation is that pairwise image alignment and joint image alignment are decoupled, and we observed that the joint alignment stage does not perfectly fit the pairwise alignment results. This issue can be addressed by performing joint alignment to match the features of the pixels. 

The third limitation is the computational cost, where stages II to IV take 10 to 15 minutes on a standard desktop. This is about two times the cost of stage I. In the future, we plan to accelerate view alignment using neural networks. 

%% file: 07_conclusions.tex
\section{Conclusions}

This paper introduces an optimization procedure to generate a textured 3D mesh from pre-trained text-to-image models. We study three types of multi-view inconsistency patterns among text-to-image output, which are used to develop our four-step optimization procedure. We show the design of view generation, view selection, view alignment, and view stitching. In particular, alternating between view alignment and view stitching demonstrates significant quality enhancement. The experimental results show that our approach outperforms state-of-the-art text-to-textured-mesh approaches both qualitatively and quantitatively. 

There are ample venues for future research. First, unlike many existing approaches that are trained end-to-end, our approach can be considered as a post-processing step. A natural question is whether it is possible to integrate different stages together into a single neural network to optimize its performance. Another direction is to combine pre-trained text-to-image models with training data offered by textured 3D models. Finally, controlling the textured mesh using latent parameters is another promising direction to reduce the uncertainties in text prompt specifications.